\documentclass[pdflatex,sn-mathphys-num,colorlinks=true,urlcolor=blue,linkcolor=blue]{sn-jnl}

\usepackage{graphicx}
\usepackage{multirow}
\usepackage{amsmath,amssymb,amsfonts}
\usepackage{amsthm}
\usepackage{mathrsfs}
\usepackage[title]{appendix}
\usepackage{xcolor}
\usepackage{textcomp}
\usepackage{manyfoot}
\usepackage{array}
\usepackage{booktabs} 
\usepackage{algorithm}
\usepackage{algorithmicx}
\usepackage{float}
\usepackage{algpseudocode}
\usepackage{listings}
\usepackage{tabularx}
\newcolumntype{Y}{>{\centering\arraybackslash}X}
\usepackage{placeins}
\usepackage{colortbl}
\usepackage{booktabs}
\definecolor{okc}{HTML}{DDF1DD} 
\definecolor{noc}{HTML}{F7DADA} 
\definecolor{unk}{HTML}{ECECEC} 

\newcommand{\ok}[1]{\cellcolor{okc}#1}
\newcommand{\no}[1]{\cellcolor{noc}#1}

\hypersetup{
  colorlinks=true,
  urlcolor=blue,
  linkcolor=blue
}

\theoremstyle{thmstyleone}%

\theoremstyle{thmstyletwo}%

\theoremstyle{thmstylethree}%

\begin{document}

\title[Article Title]{\textbf{EndoS$^3$}:\textbf{Endo}scopic \textbf{S}emantic \textbf{S}platting \textbf{S}LAM}


\author*[1]{\fnm{Mona} \sur{Sheikh Zeinoddin}}\email{mona.zeinoddin.22@ucl.ac.uk, e.mazomenos@ucl.ac.uk}
\equalcont{These authors contributed equally to this work.}
\author[1]{\fnm{Benjamin} \sur{Wright}}
\equalcont{These authors contributed equally to this work.}

\author[2]{\fnm{Zafer} \sur{Tandogdu}}

\author[2]{\fnm{Greg} \sur{Shaw}}

\author[1]{\fnm{Evangelos} \sur{Mazomenos}}
\author[1]{\fnm{Danail} \sur{Stoyanov}}

\affil[1]{\orgdiv{Hawkes Institute}, \orgname{University College London}, \country{UK}}
\affil[2]{\orgdiv{Dept of Urology}, \orgname{University College London Hospitals}, \country{UK}}


\abstract{\textbf{Purpose}: Despite advances in 3D surgical scene understanding, semantic simultaneous localisation and mapping (SLAM) remains under-explored for two key reasons. First, it requires solving three challenging tasks in parallel: camera pose tracking, high-fidelity deformable reconstruction, and semantic mapping, making progress particularly demanding. Second, the scarcity of datasets providing the necessary ground truth to evaluate all three tasks comprehensively under conditions that closely reflect real surgical procedures environments. 

\textbf{Method}:
We propose \textbf{EndoS$^3$}, a Gaussian Splatting-based semantic SLAM framework that uses instance masks to guide all three tasks. An instance-level motion decomposition module separates static and dynamic instances by evaluating residual flow after compensating for camera motion, accounting for flow noise and comparing motion across neighbouring instances. The camera is tracked against the static Gaussians, while an instance-aware warp field gives each dynamic instance its own
control nodes and deformation network. A semantic node allocation module redistributes a fixed node budget across instances according to their size and non-rigidity. We also introduce \textbf{CRCD++}, an \textit{ex-vivo} Cholecystectomy dataset with RGB-D frames, camera pose and instance annotations for full semantic SLAM evaluation.

\textbf{Results}: \textbf{EndoS$^3$} surpasses all state-of-the-art general-purpose and surgical semantic SLAM methods in tracking, reconstruction and semantic mapping on the \textbf{CRCD++} and SemSup datasets.

\textbf{Conclusion}: Semantic-aware motion and deformation modelling is vital to surgical semantic SLAM. \textbf{EndoS$^3$} and \textbf{CRCD++} offer a method and benchmark for improved geometric and semantic understanding in endoscopy.}




\maketitle
\section{Introduction}

Spatial understanding of the endoscopic field supports surgical navigation
and context-aware decision-making~\cite{lin2023semantic,acar2026perseus}.
Reaching it requires complementary geometric and semantic information: geometry
describes the spatial relationships between the camera, anatomy, and
instruments, while semantics identifies them and analyses their interactions. Surgical semantic simultaneous
localisation and mapping (SLAM) achieves this understanding through jointly performing camera
pose tracking, scene reconstruction, and semantic mapping, though it suffers from several key challenges.

First, image motion is inherently ambiguous: pixel displacements may arise from camera motion, tissue deformation, or independently moving instruments~\cite{hayoz2023robust}. Misattributing scene motion to
the camera corrupts pose estimates, while the opposite degrades the
reconstruction. Previous approaches use geometric cues to decompose the sources of motion and limit the
influence of tissue deformation on camera pose estimation~\cite{hayoz2023robust},
yet the role of instance-level semantics in motion decomposition remains
under-explored, even though it can provide valuable information by associating pixels with entities that exhibit distinct motion characteristics.

Second, surgical scenes exhibit heterogeneous, time-varying deformations: Rigid instruments move quickly and
independently of the tissue they manipulate, while organs deform slowly and largely coherently. Deformation models must capture these differences and adapt to the content currently in view. Current approaches, however, model the motion of the entire scene with a single warp field, driven by a shared
set of control nodes~\cite{huang2024sc,hayoz2024online,li20254d}, which can blend instrument and tissue motion near their contact points. Third, a fixed set of control nodes~\cite{li20254d} also limit
adaptation to instruments entering or leaving the view and regions of
greater tissue deformation. Existing methods primarily use semantics for map labelling
and data association~\cite{shan2024dds,zhu2024sni,li2024sgs,zhu2025semgauss},
while their use in motion decomposition and deformation modelling remains unexplored.
Moreover, progress is further limited by the scarcity of evaluation data.
Existing datasets lack joint ground truth for
camera poses, depth, and segmentation~\cite{lin2023semantic,hayoz2023robust}. These annotations are needed to evaluate tracking, reconstruction, and
semantic mapping jointly and determine how improvements in one task affect the others.

We propose \textbf{EndoS$^3$}, a semantic SLAM framework that uses instance semantics to guide motion decomposition for camera tracking and deformable semantic mapping. We also present \textbf{CRCD++}, a cholecystectomy dataset with camera poses, depth, and instance annotations for a complete evaluation of semantic SLAM methods. Our code and data are at: \url{https://bwright000.github.io/EndoS3/}. Our main contributions are:

\begin{itemize}

    \item \textbf{Instance-level motion decomposition}, which separates static and dynamic instances by investigating camera-compensated flow residuals, their signal-to-noise ratios, and motion differences across instance boundaries.

    \item \textbf{Instance-aware warp field}, which assigns every instance its own control nodes and deformation network
    and restricts motion propagation to within that instance.

    \item \textbf{Semantic node allocation}, which redistributes a fixed node budget across instances at every frame
    according to their size and non-rigidity.

    \item \textbf{CRCD++}, an \textit{ex-vivo} Cholecystectomy dataset with all of the necessary ground truth needed to perform an in-depth evaluation of surgical semantic SLAM methods.
\end{itemize}

\section{Related Work}
\textbf{General-purpose semantic SLAM:} Dense semantic SLAM builds the map using implicit fields or explicit primitives. SNI-SLAM~\cite{zhu2024sni} fuses appearance, geometry and semantic features in a neural implicit map through cross-attention; SGS-SLAM~\cite{li2024sgs} jointly optimises the appearance, geometric and semantic features of a Gaussian map and uses semantics to select keyframes; and SemGauss-SLAM~\cite{zhu2025semgauss} embeds semantic features in the Gaussians and enforces consistency through semantic-informed bundle adjustment. All three regularise a static map and are evaluated on indoor scenes~\cite{straub2019replica}. None model deformation or dynamics, so in scenes that feature either, they cannot disentangle scene from camera motion, typically using the latter to explain the former.
\noindent
\textbf{Surgical semantic SLAM:} Among sparse methods, Semantic-SuPer~\cite{lin2023semantic} builds a surfel map with a sparse embedded deformation graph and uses semantic masks within tissue tracking to pull surfels into their semantic region, but assumes a stationary camera. PERSEUS~\cite{acar2026perseus} combines learned segmentation and monocular depth with a learning-based SLAM backend to label a point cloud. It neither produces high-fidelity renderings nor models deformation, and is evaluated only on static sequences. Among dense methods, DDS-SLAM~\cite{shan2024dds} models deformation as a hash-grid canonical scene with a time-dependent neural field and uses a semantic distance loss. Deformation is supervised only through rendering, and tracking does not separate static from moving regions, leaving camera and tissue motion entangled. It is evaluated on deforming scenes~\cite{hayoz2023robust,lin2023semantic} and outputs no semantic map; as they are used in its optimisation solely.

None of these methods use semantics to disentangle scene motion from camera motion. \textbf{EndoS$^3$} uses a Gaussian Splatting framework where instance semantics guides motion decomposition: the camera is tracked against the static instances only, separating camera from scene motion, while an instance-aware warp field models moving organs and instruments individually and allows each to deform independently.
\begin{table}[b]
\renewcommand{\tabularxcolumn}[1]{m{#1}}
\caption{Overview of Semantic datasets. Sem.: semantic classes ($^{s}$sparse; $^{t}$tools only; $^{nt}$no tools); Seq.\ len.: sequence length (frames); GT/Pseudo-GT: measured/stereo-estimated depth. Green: met; red: not met.}
\label{tab:surgical_datasets}
\setlength{\tabcolsep}{1.5pt}
\begin{tabularx}{\textwidth}{l|YYYYYYY|>{\centering\arraybackslash}m{1.85cm}}
\hline
 & Cholec-\newline Seg8k~\cite{hong2020cholecseg8k} & Endo-\newline Vis18~\cite{allan2020endovis18} & Endo-\newline scapes~\cite{mascagni2025endoscapes} & Semantic-\newline SuPer~\cite{lin2023semantic} & Stereo-\newline MIS~\cite{hayoz2023robust} & D4D\newline \cite{docea2026dresden} & eCRCD\newline \cite{oh2025expanded} & \textbf{CRCD++}\\
\hline
Poses & \no{No} & \no{No} & \no{No} & \no{No} & \ok{Yes} & \ok{Yes} & \ok{Yes} & \ok{\textbf{Yes}}\\
Depth & \no{No} & \no{No} & \no{No} & \no{No} & \no{No} & \ok{GT} & \no{No} & \ok{\textbf{\mbox{Pseudo-GT}}}\\
Sem. & \ok{13} & \no{12$^{s}$} & \no{6$^{s}$} & \ok{3} & \no{1$^{t}$} & \no{1$^{t}$} & \no{2$^{nt}$} & \ok{\textbf{4}}\\
Seq.\ len. & \no{80} & \no{Sparse} & \no{Sparse} & \ok{150} & \ok{1.4--9.6k} & \ok{$\sim$3k} & \no{120} & \ok{\textbf{0.2--2.1k}}\\
\hline
Frames & 8,080 & 2,384 & 493 & 600 & $\sim$58.7k & 300k+ & -- & \textbf{14,688}\\
\hline
\end{tabularx}
\end{table}
\noindent
\newline
\textbf{Surgical benchmarks:} Semantic SLAM evaluation requires RGB images, depth, camera poses, and segmentation annotations. Existing datasets~\cite{hong2020cholecseg8k,allan2020endovis18,mascagni2025endoscapes} offer sparse segmentations or lack poses. StereoMIS~\cite{hayoz2023robust} and D4D~\cite{docea2026dresden} provide poses but segment only tools; D4D also provides measured depth. eCRCD~\cite{oh2025expanded} annotates two tissue classes but lacks tool masks and propagates labels from one frame per clip. \textbf{CRCD++} alone meets all four requirements. Unlike the general abdominal scenes in StereoMIS and D4D, \textbf{CRCD++} provides clinically meaningful organ annotations specific to Cholecystectomy (Table~\ref{tab:surgical_datasets}).

\section{Methods}

\begin{figure}[t]
  \centering
  \includegraphics[width=1\textwidth]{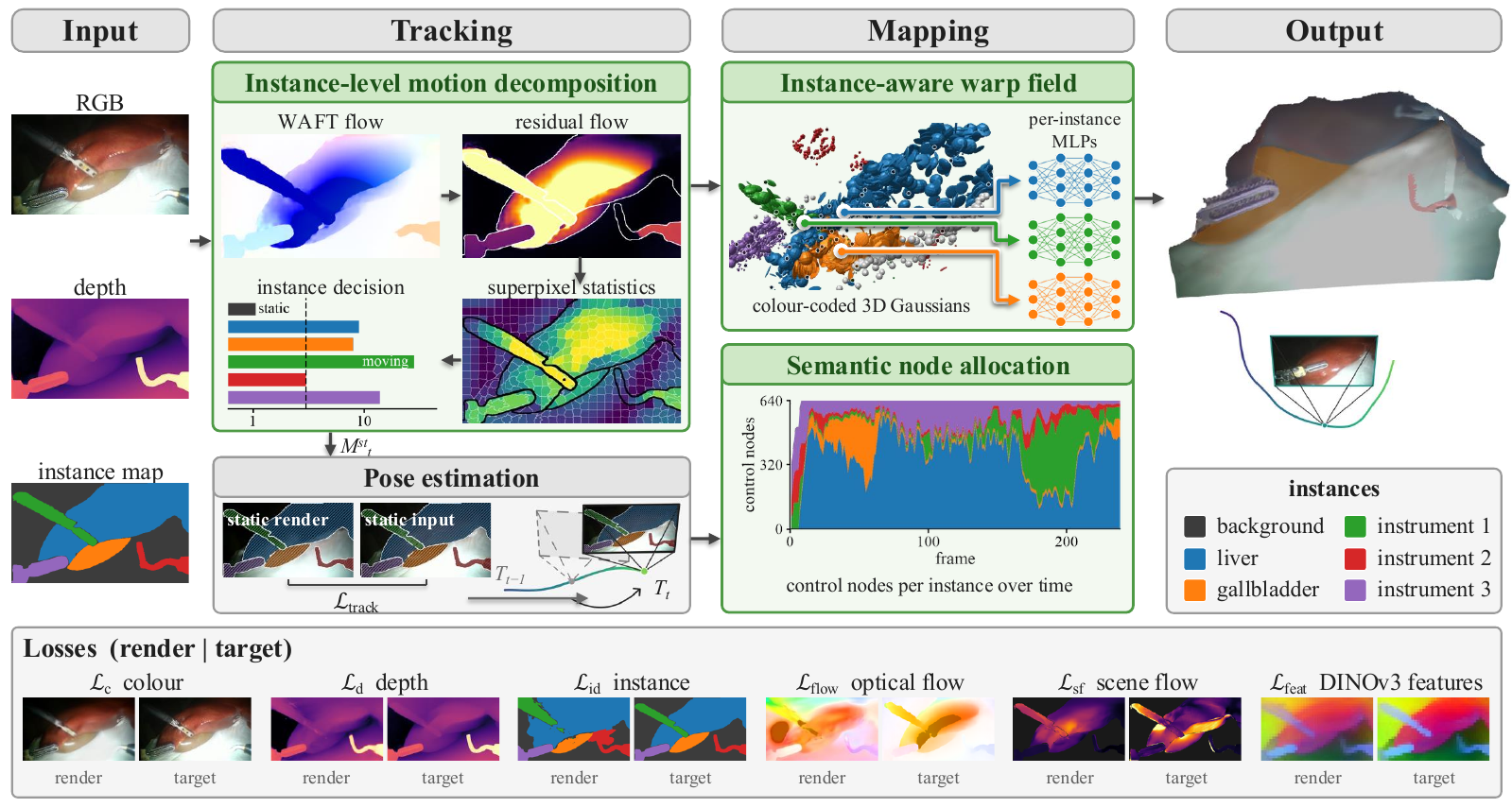}
  \caption{Overview of our framework - EndoS$^3$ uses semantics in both tracking and mapping.}
  \label{fig:main}
\end{figure}

As shown in Fig.~\ref{fig:main}, given an RGB-D sequence and per-frame instance segmentation masks $\{\mathbf{C}_t,\mathbf{D}_t\, \mathbf{S}_t\}$, EndoS$^3$ adopts a 3D semantic Gaussian Splatting scene representation~\cite{zhu2025semgauss}. Our Instance-level motion decomposition module, separates the Gaussians into (semi-) static and dynamic. Tracking is then performed on the static part of the scene, while mapping utilises our Instance-aware warp field module to model deformations in the scene. This module operates on a fixed set of control nodes, distributed between instance regions using our Semantic node allocation module, and propagates the estimated deformation to all dynamic Gaussians in an instance region. 



\subsection{Instance-level motion decomposition.}

The optical flow $\mathbf{F}_{t'\to t}$ from an earlier frame $t'=t-5$ to the current frame $t$ is estimated by
zero-shot WAFT~\cite{wang2026waft}. Each pixel $\mathbf{u}$ of
frame $t'$ with depth $\mathbf{D}_{t'}$ is back-projected to $\mathbf{X}(\mathbf{u})=\mathbf{D}_{t'}(\mathbf{u})\,\mathbf{K}^{-1}\tilde{\mathbf{u}}$,
with $\mathbf{K}$ the camera intrinsics and $\tilde{\mathbf{u}}$ the homogeneous pixel coordinates. The optical flow provides
its location $\mathbf{u}+\mathbf{F}_{t'\to t}(\mathbf{u})$ in frame $t$. Leveraging these 3D--2D correspondences,
PnP-RANSAC estimates the relative camera rotation $\hat{\mathbf{R}}$ and
translation $\hat{\mathbf{t}}$ between $t'$ and $t$, with independently moving pixels rejected as outliers. The noisy residual flow is:
\begin{equation}
  \mathbf{r}(\mathbf{u})=\mathbf{F}_{t'\to t}(\mathbf{u})
  -\Bigl(\pi\bigl(\hat{\mathbf{R}}\,\mathbf{X}(\mathbf{u})+\hat{\mathbf{t}}\bigr)-\mathbf{u}\Bigr),
  \label{eq:residual}
\end{equation}
where $\pi$ is the projection function. To obtain robust region-level evidence, the residuals are aggregated in SLIC superpixels split
along the instance boundaries, so that each superpixel $P$ lies within a single instance. Each $P$ is summarised by its
median residual $\bar{\mathbf{r}}_P$ and its median forward--backward flow error $\bar{e}_P$, which estimates the local
flow uncertainty. An instance $\ell$, covered by the superpixels $\mathcal{P}_\ell$, is then tested with two
complementary statistics:
\begin{equation}
  z^{A}_{\ell}=\operatorname*{med}_{P\in\mathcal{P}_{\ell}}\frac{\|\bar{\mathbf{r}}_P\|_2}{\bar{e}_P},\qquad
  z^{B}_{\ell\ell'}=\frac{\|\mathbf{m}_{\ell}-\mathbf{m}_{\ell'}\|_2}{\sqrt{s_{\ell}^2/n_{\ell}+s_{\ell'}^2/n_{\ell'}}} .
  \label{eq:keys}
\end{equation}
$z^A_\ell$ measures residual motion relative to flow noise, while
$z^B_{\ell\ell'}$ measures motion differences across the boundary with
neighbouring instance $\ell'$. For boundary superpixels
$\mathcal{Q}_{\ell\ell'}\subseteq\mathcal{P}_\ell$, we define the median
residual, robust scale estimate, and superpixel count:
\begin{equation}
\mathbf{m}_\ell=\operatorname*{med}_{P\in\mathcal{Q}_{\ell\ell'}}\bar{\mathbf{r}}_P,
\qquad
s_\ell={MAD}_{P\in\mathcal{Q}_{\ell\ell'}}\bigl\|\bar{\mathbf{r}}_P-\mathbf{m}_\ell\bigr\|,
\qquad
n_\ell=|\mathcal{Q}_{\ell\ell'}|,
\end{equation}
The same quantities are computed on the neighbouring side.
The tests are complementary: absolute residuals may reflect pose errors,
whereas boundary differences suppress shared errors but cannot determine
which instance moved. An instance is classified as dynamic if both
statistics exceed $\varepsilon$ for at least one neighbour and its
residual is larger than the neighbour's. We use one threshold across
both tests and all sequences. The module outputs the static mask
$M^{\mathrm{st}}_t$ for tracking, and its complement
$M^{\mathrm{dyn}}_t = \sim M^{\mathrm{st}}_t$, the bidirectional flow, and the per-instance $s_\ell$ as a measure of non-rigidity for mapping.

\subsection{Tracking}
The pose $\mathbf{T}_t$ is optimised from the previous pose against the the static part $M^{\mathrm{st}}_t$~\cite{li20254d}, using the colour  $\mathcal{L}_{c}$, depth $\mathcal{L}_{d}$ and instance segmentation $\mathcal{L}_{id}$ losses~\cite{shan2024dds}:
\begin{equation}
  \begin{split}
    \mathcal{L}_{\mathrm{track}} ={}& \lambda_{\mathrm{c}}\,\mathcal{L}_{c} + (1-\lambda_{\mathrm{d}})\,\mathcal{L}_{d} + \lambda_{\mathrm{id}}\,\mathcal{L}_{\mathrm{id}}
  \end{split}
  \label{eq:track}
\end{equation}

\subsection{Mapping}
\label{sec:mapping}
\noindent
\textbf{Instance-aware Warp field}:
We model the motion of the dynamic Gaussians with a sparse set of control nodes~\cite{huang2024sc}, partitioned by instance, so
that each instance deforms independently. The dynamic Gaussians are kept in a time-independent canonical space, and the
nodes define a warp field that maps the canonical space to their state at time $t$. Each node $k$ has a canonical position $\mathbf{p}_k$
and an instance label $\ell_k$, and the deformation MLP $\Psi_{\ell_k}$ of its instance predicts its deformation at time $t$:
\begin{equation}
  (\delta\mathbf{x}_k,\delta\mathbf{q}_k,\,\delta\mathbf{s}_k)=\Psi_{\ell_k}\bigl(\gamma(\mathbf{p}_k),\,\gamma(t)\bigr),
  \label{eq:warp}
\end{equation}
where $\delta\mathbf{x}_k\in\mathbb{R}^{3}$,
$\delta\mathbf{q}_k\in\mathbb{R}^{4}$ and $\delta\mathbf{s}_k\in\mathbb{R}^{3}$ are offsets to the position, rotation and
scale of the Gaussians it drives, and $\gamma$ denotes the frequency-based positional encoding~\cite{mildenhall2020nerf}.
Linear blend skinning~\cite{huang2024sc} propagates these outputs to each dynamic Gaussian $i$ from its neighbourhood
$\mathcal{N}_i$, the $K$ nearest nodes in its own instance $\ell_i$. 
\newline
\newline
\noindent
\textbf{Semantic node allocation}: A fixed budget of $B$ nodes is redistributed across instances every frame. Instances
whose dynamic area $|\Omega^t_\ell|$, i.e.\ their number of pixels in the dynamic mask $M^{\mathrm{dyn}}_t$, exceeds a
minimum, form the active set $\mathcal{A}_t$. All other instances keep at most $n_{\mathrm{abs}}$ nodes to preserve their
past motion, and the remaining budget $B_{\mathrm{free}}$ is shared among the active instances with targets:
\begin{equation}
B_\ell=\max\Bigl(n_{\min},\,\operatorname{round}\Bigl(B_{\mathrm{free}}\,
\frac{w_\ell}{\sum_{\ell'\in\mathcal{A}_t}w_{\ell'}}\Bigr)\Bigr),
\qquad
w_\ell=\sqrt{|\Omega^t_\ell|}\,(1+s_\ell),
\label{eq:alloc}
\end{equation}
where $n_{\min}$ is the minimum allocation and $s_\ell$ the instance non-rigidity: the square root limits the
advantage of large instances, while $s_\ell$ favours deforming ones. Deficits are filled first from unused nodes, then
from instances above their target, which give up their least-supported nodes that are the nearest node of
the fewest Gaussians.
\newline
\newline
\noindent
\textbf{Optimisation:} The mapping objective calculated on the dynamic part $M^{\mathrm{dyn}}_t$ is:
\begin{equation}
  \begin{split}
    \mathcal{L}_{\mathrm{map}} ={}&
    \lambda\,\mathcal{L}_{c} + (1-\lambda)\,\mathcal{L}_{d}
    + \lambda_{\mathrm{f}}\,\mathcal{L}_{\mathrm{flow}} + \lambda_{\mathrm{sf}}\,\mathcal{L}_{\mathrm{sf}} + \lambda_{\mathrm{id}}\,\mathcal{L}_{\mathrm{id}} + \lambda_{\mathrm{feat}}\,\mathcal{L}_{\mathrm{feat}}
    + \mathcal{L}_{\mathrm{reg}},
  \end{split}
  \label{eq:map}
\end{equation}
where $\mathcal{L}_{c}$ , $\mathcal{L}_{d}$ and $\mathcal{L}_{\mathrm{id}}$ are the photometric, depth and instance segmentation losses~\cite{li20254d};
$\mathcal{L}_{\mathrm{flow}}$ and $\mathcal{L}_{\mathrm{sf}}$ are used following~\cite{li20254d,wu2025endoflow, wang2026flow4dgs} for optical
flow and scene
flow; $\mathcal{L}_{\mathrm{feat}}$ compares  features~\cite{simeoni2025dinov3} of the rendered and input images, from the frozen DINO V3 encoder of WAFT following~\cite{zhu2024sni}; and $\mathcal{L}_{\mathrm{reg}}$
comprises the isotropy and ARAP terms~\cite{huang2024sc,wang2026flow4dgs}. Following~\cite{wang2026flow4dgs},
dynamic Gaussians have a temporal GMM opacity, which we gate by instance. Loss weights and further details are given in the supplementary material.

\label{sec:methods}
\subsection{CRCD++}
\label{sec:crcdpp}
\begin{figure}[h]
  \centering
  \includegraphics[width=0.79\textwidth]{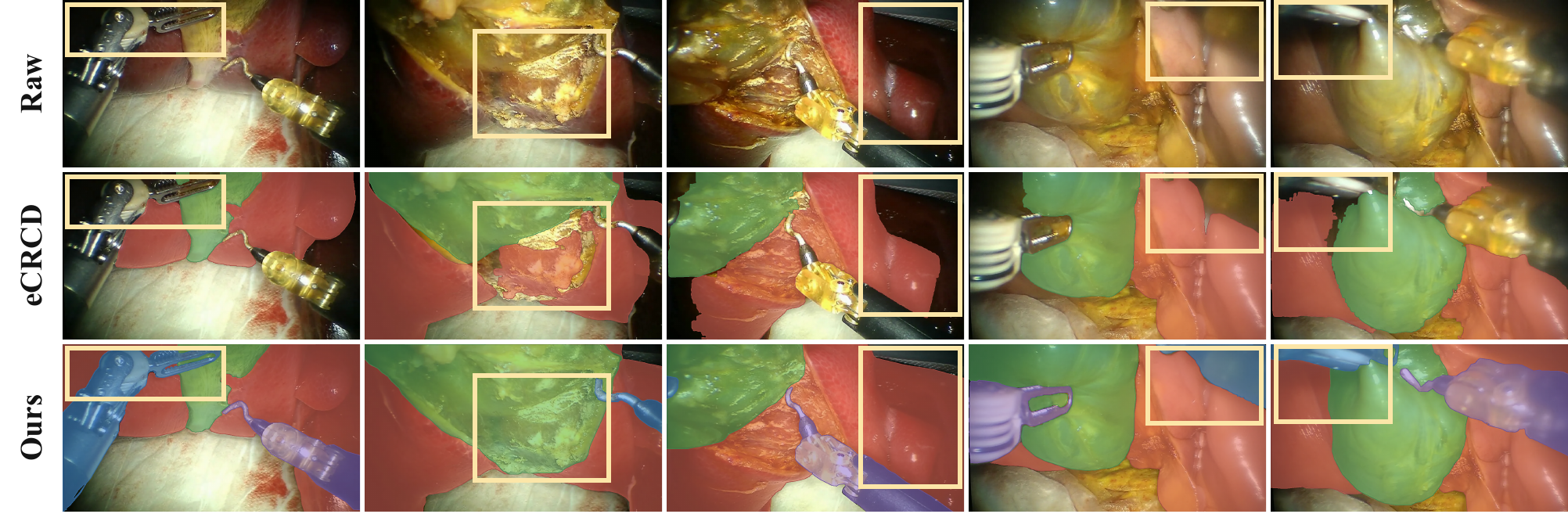}
  \caption{Semantic annotations. Input | eCRCD Track Anything | Our manual annotations.}
  \label{fig:annotation}
\end{figure}

The expanded Comprehensive Robotic Cholecystectomy Dataset (eCRCD)~\cite{oh2025expanded} records 18 ex-vivo porcine Cholecystectomies by seven surgeons on a dVRK-controlled da~Vinci, with stereo video and camera kinematics. Its tissue masks are propagated by Track Anything~\cite{yang2023track} from one hand-labelled frame per 120-frame clip and therefore, degrade under occlusion, camera motion and deformation. Finally, eCRCD does not provide dense tool masks, and only provides keypoints for them. In order to address the limitations of eCRCD, we refine a subset of 20 sequences (200--2108 frames; 14{,}688 stereo pairs), and augment every frame as follows. Dense masks for Four semantic classes (background, liver, gallbladder, tool) are produced by a human annotator. Each tool carries its own instance ID(Fig.~\ref{fig:annotation}).
Depth is estimated on every rectified pair with zero-shot FoundationStereo (FS)~\cite{wen2025foundationstereo}.
\section{Experiments}
\label{sec:experiments}
\textbf{Datasets}:
Five \noindent\textbf{CRCD++} sequences: C1, C2, E3, C3 and G3 and Semantic-Super (\textbf{SemSup})~\cite{lin2023semantic} test sequences with the pseudo GT depth we produce using FS.

\noindent
\textbf{Implementation and Evaluation}: We compare against DDS-SLAM~\cite{shan2024dds}, SNI-SLAM~\cite{zhu2024sni}, SGS-SLAM~\cite{li2024sgs}, SemGauss-SLAM~\cite{zhu2025semgauss}, Semantic-SuPer~\cite{lin2023semantic} and PERSEUS~\cite{acar2026perseus}, using their public repositories. Semantic methods receive each dataset's semantic maps as input (the ground-truth labels on CRCD++ and the released DeepLabV3+ masks on Semantic-SuPer). All experiments use one NVIDIA A100 (40\,GB). In evaluation, We use ATE-RMSE for tracking, PSNR, SSIM and LPIPS for mapping. On Semsup, tissue tracking is evaluated by reprojection error using the tracked sparse pins. For additional visualisations, examples of complex scenarios, and a more detailed performance review, please refer to the supplementary material.

\section{Results}
\label{sec:results}
\begin{figure}[t]
  \centering
  \includegraphics[width=0.84\textwidth]{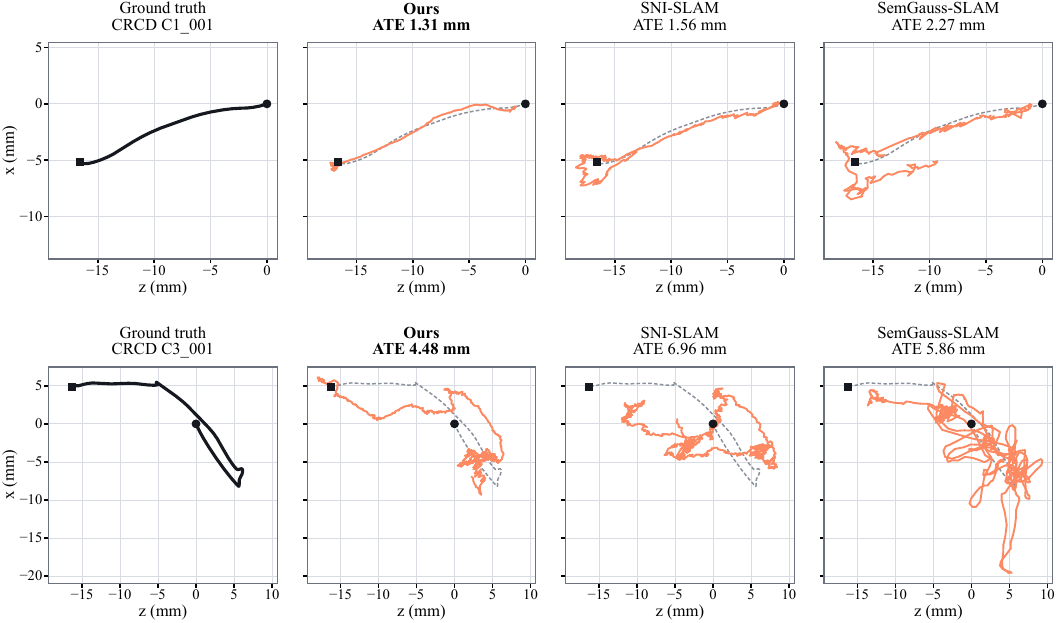}
  \caption{Our camera tracking results on CRCD++ C1 and C3 - our Instance-level motion decomposition allows EndoS$^3$ to produce much smoother trajectories.}
  \label{fig:trajectory}
\end{figure}

\begin{figure}[!]
  \centering
  \includegraphics[width=0.99\textwidth]{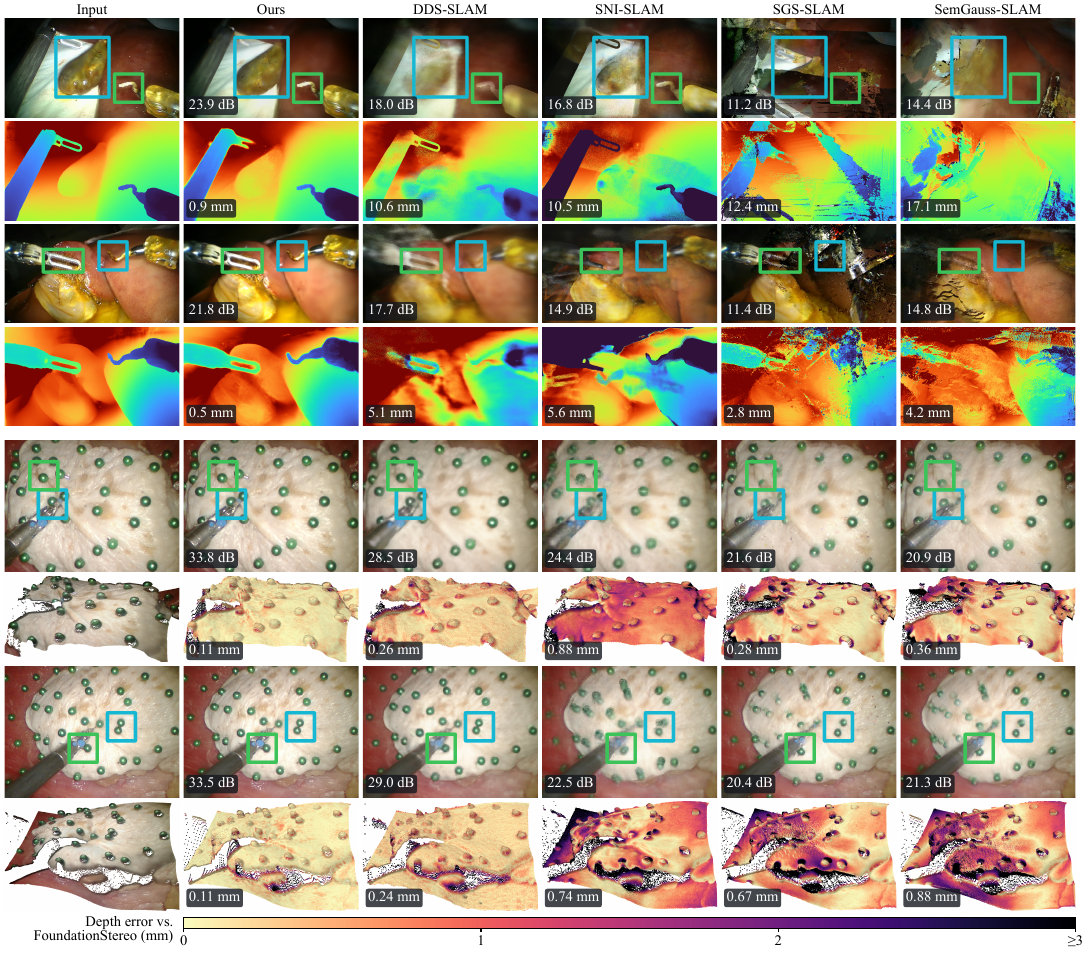}
  \caption{CRCD++ and SemSuP: Rendered colour map, CRCD++: Rendered depth map, SemSuP: Depth rendered as a pointcloud error heatmap.}
  \label{fig:qualitative}
\end{figure}

\textbf{Tracking on CRCD++:} Table~\ref{tab:geom} shows EndoS$^3$ achieves the best ATE result in 3/5 sequences and the lowest average, with an improvement of 17\% over the second best model. Fig.~\ref{fig:trajectory} demonstrates that the benchmarked models cannot separate camera motion from tissue deformation, and absorb deformation into the camera pose updates, producing jittery trajectories compared to our smoother predictions, the result of using our Instance-level motion decomposition module.
\newline
\noindent\textbf{Reconstruction and Depth:} Table~\ref{tab:render} shows we achieve an average improvement of 2 dB and 0.81 dB over the second best model on CRCD++ and SemSup respectively. In depth estimation, Table~\ref{tab:geom} shows that on both datasets, EndoS$^3$ leads every tissue tracking (Pins) and depth metric. This is also visible in Fig. \ref{fig:qualitative}.
\newline
\begin{table}[t]
\centering
\caption{\textbf{Tracking, depth and semantics.} ATE-RMSE and depth error in mm; mIoU in \%; SemSuP: pin reprojection error in px$\downarrow$. ``--'': not available.}
\label{tab:geom}
\setlength{\tabcolsep}{2pt}
\begin{tabularx}{\textwidth}{l|r|Y|Y|Y|Y|Y|Y|c}
\hline
\multirow{2}{*}{Method} & \multirow{2}{*}{Metric} & \multicolumn{6}{c|}{CRCD++} & Sem.-SuPer\\
\cmidrule{3-9}
 & & C1 & C2 & C3 & G3 & E3 & Avg. & Pins\\
\hline
\multirow{2}{*}{DDS-SLAM~\cite{shan2024dds}} & ATE$\downarrow$ & 1.78 & 3.36 & 6.59 & 4.83 & \textbf{5.13} & 4.34 & \multirow{2}{*}{9.21}\\
 & Depth$\downarrow$ & 11.1 & 19.9 & 13.4 & 8.0 & \underline{14.8} & 13.45 & \\
\hline
\multirow{3}{*}{SNI-SLAM~\cite{zhu2024sni}} & ATE$\downarrow$ & \underline{1.56} & \textbf{1.48} & 6.96 & 2.60 & 5.75 & \underline{3.67} & \multirow{3}{*}{9.70}\\
 & Depth$\downarrow$ & 5.1 & 17.9 & 14.3 & 8.8 & 17.6 & 12.72 & \\
 & mIoU$\uparrow$ & 63.6 & 52.1 & 60.2 & 32.6 & 41.5 & 50.0 & \\
\hline
\multirow{3}{*}{SGS-SLAM~\cite{li2024sgs}} & ATE$\downarrow$ & 2.13 & 5.34 & 8.54 & 4.24 & \underline{5.21} & 5.09 & \multirow{3}{*}{\underline{9.04}}\\
 & Depth$\downarrow$ & \underline{3.1} & \underline{16.8} & \underline{6.2} & \underline{2.3} & 21.7 & \underline{10.04} & \\
 & mIoU$\uparrow$ & \underline{86.3} & \underline{53.9} & \underline{62.2} & \textbf{52.8} & 42.2 & \underline{59.5} & \\
\hline
\multirow{3}{*}{SemGauss-SLAM~\cite{zhu2025semgauss}} & ATE$\downarrow$ & 2.27 & 5.00 & \underline{5.86} & \underline{1.87} & 5.41 & 4.08 & \multirow{3}{*}{9.59}\\
 & Depth$\downarrow$ & 3.2 & 18.5 & 8.4 & 3.2 & 26.0 & 11.86 & \\
 & mIoU$\uparrow$ & 84.8 & 41.2 & 48.2 & 48.9 & \underline{48.4} & 54.3 & \\
\hline
\multirow{2}{*}{PERSEUS~\cite{acar2026perseus}} & ATE$\downarrow$ & 2.90 & 5.29 & 9.90 & 2.92 & 5.82 & 5.37 & \multirow{2}{*}{10.16}\\
 & Depth$\downarrow$ & 20.7 & 114.9 & 75.0 & 6.5 & 302.4 & 103.90 & \\
\hline
Semantic-SuPer~\cite{lin2023semantic} & Depth$\downarrow$ & 10.2 & -- & 16.1 & 3.1 & 35.3 & -- & 23.17\\
\hline
\multirow{3}{*}{\textbf{Ours}} & ATE$\downarrow$ & \textbf{1.31} & \underline{2.24} & \textbf{4.48} & \textbf{1.80} & 5.29 & \textbf{3.03} & \multirow{3}{*}{\textbf{8.20}}\\
 & Depth$\downarrow$ & \textbf{2.2} & \textbf{4.6} & \textbf{3.1} & \textbf{1.8} & \textbf{3.1} & \textbf{2.96} & \\
 & mIoU$\uparrow$ & \textbf{91.5} & \textbf{78.9} & \textbf{84.3} & \underline{51.9} & \textbf{81.9} & \textbf{77.7} & \\
\hline
\end{tabularx}
\end{table}
\begin{table}[t]
\centering
\caption{\textbf{Rendering quality on CRCD++ and Semantic-SuPer.} PERSEUS has no renderer.}
\label{tab:render}
\setlength{\tabcolsep}{2pt}
\begin{tabularx}{\textwidth}{l|r|Y|Y|Y|Y|Y|Y|c}
\hline
\multirow{2}{*}{Method} & \multirow{2}{*}{Metric} & \multicolumn{6}{c|}{CRCD++} & Sem.-SuPer\\
\cmidrule{3-9}
 & & C1 & C2 & C3 & G3 & E3 & Avg. & Avg.\\
\hline
\multirow{3}{*}{DDS-SLAM~\cite{shan2024dds}} & PSNR$\uparrow$ & \underline{22.19} & \underline{19.52} & \underline{21.68} & \textbf{22.90} & \underline{21.47} & \underline{21.55} & \underline{30.81}\\
 & SSIM$\uparrow$ & 0.667 & \underline{0.583} & 0.596 & 0.580 & \underline{0.639} & 0.613 & \underline{0.792}\\
 & LPIPS$\downarrow$ & 0.669 & 0.720 & 0.717 & 0.734 & \underline{0.644} & 0.697 & \underline{0.248}\\
\hline
\multirow{3}{*}{SNI-SLAM~\cite{zhu2024sni}} & PSNR$\uparrow$ & 19.32 & 16.09 & 16.60 & 16.39 & 18.61 & 17.40 & 22.47\\
 & SSIM$\uparrow$ & 0.694 & 0.552 & 0.631 & 0.703 & 0.577 & 0.631 & 0.660\\
 & LPIPS$\downarrow$ & 0.584 & 0.701 & 0.640 & 0.615 & 0.660 & 0.640 & 0.448\\
\hline
\multirow{3}{*}{SGS-SLAM~\cite{li2024sgs}} & PSNR$\uparrow$ & 20.79 & 13.81 & 18.30 & 18.44 & 12.13 & 16.69 & 22.72\\
 & SSIM$\uparrow$ & \underline{0.782} & 0.524 & \underline{0.780} & \underline{0.810} & 0.486 & \underline{0.676} & 0.721\\
 & LPIPS$\downarrow$ & \underline{0.342} & \underline{0.594} & \underline{0.460} & \underline{0.494} & 0.646 & \underline{0.507} & 0.333\\
\hline
\multirow{3}{*}{SemGauss-SLAM~\cite{zhu2025semgauss}} & PSNR$\uparrow$ & 20.21 & 15.03 & 17.05 & 17.67 & 11.95 & 16.38 & 23.01\\
 & SSIM$\uparrow$ & 0.770 & 0.563 & 0.734 & 0.779 & 0.491 & 0.667 & 0.718\\
 & LPIPS$\downarrow$ & 0.347 & 0.605 & 0.515 & 0.556 & 0.701 & 0.545 & 0.304\\
\hline
\multirow{3}{*}{Semantic-SuPer~\cite{lin2023semantic}} & PSNR$\uparrow$ & 11.02 & 7.87 & 7.74 & 8.11 & 9.90 & 8.93 & 17.12\\
 & SSIM$\uparrow$ & 0.257 & 0.058 & 0.154 & 0.135 & 0.230 & 0.167 & 0.542\\
 & LPIPS$\downarrow$ & 0.819 & 0.888 & 0.992 & 0.941 & 0.858 & 0.900 & 0.611\\
\hline
\multirow{3}{*}{\textbf{Ours}} & PSNR$\uparrow$ & \textbf{29.26} & \textbf{20.28} & \textbf{22.95} & \underline{19.12} & \textbf{26.14} & \textbf{23.55} & \textbf{31.62}\\
 & SSIM$\uparrow$ & \textbf{0.893} & \textbf{0.783} & \textbf{0.813} & \textbf{0.867} & \textbf{0.868} & \textbf{0.845} & \textbf{0.867}\\
 & LPIPS$\downarrow$ & \textbf{0.250} & \textbf{0.419} & \textbf{0.383} & \textbf{0.375} & \textbf{0.321} & \textbf{0.350} & \textbf{0.217}\\
\hline
\end{tabularx}
\end{table}

\begin{figure}[!h]
  \centering
  \includegraphics[width=0.94\textwidth]{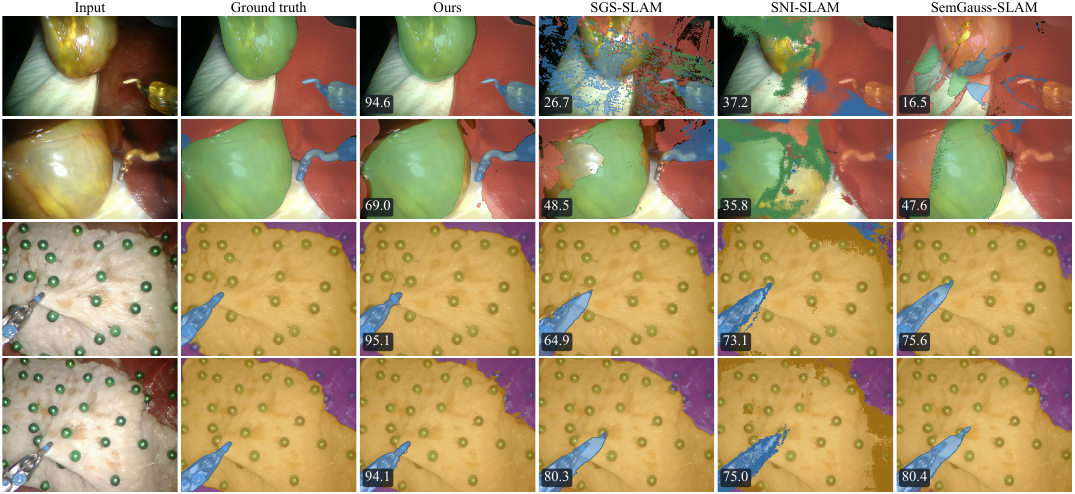}
  \caption{Semantic renders on CRCD++ and Semantic-SuPer.}
  \label{fig:semantics}
\end{figure}

\noindent\textbf{Semantics:} EndoS$^3$ achieves the highest mIoU on both datasets in Table~\ref{tab:geom}, while Fig.~\ref{fig:semantics} shows the baselines' semantic degradation on CRCD++ and SemSup.
\newline
\noindent\textbf{Ablation studies:} Table~\ref{tab:ablation} identifies motion decomposition as the most impactful module, with its removal causing the largest degradation. After that, Instance-aware deformation contributes more to depth accuracy, whereas semantic node allocation has a greater impact on rendering, tracking, and semantics. 

\begin{table}[!h]
\centering
\caption{\textbf{Ablation on CRCD++ - seq C4.}
MD: motion decomposition (otherwise only instruments are dynamic);
IW: instance-aware warp field (otherwise one shared MLP);
NA: semantic node allocation (otherwise a fixed node set).}
\label{tab:ablation}
\setlength{\tabcolsep}{2pt}
\begin{tabularx}{\textwidth}{ccc|YYYY}
\hline
MD & IW & NA & PSNR$\uparrow$ & Depth$\downarrow$
& ATE$\downarrow$ & mIoU$\uparrow$ \\
\hline
$\times$ & $\checkmark$ & $\checkmark$
& 22.18 & 5.85 & 2.91 & 52.8 \\
$\checkmark$ & $\times$ & $\checkmark$
& \underline{23.28} & 4.62 & \underline{2.55} & \underline{71.3} \\
$\checkmark$ & $\checkmark$ & $\times$
& 23.08 & \underline{4.43} & 2.83 & 70.9 \\
\hline
$\checkmark$ & $\checkmark$ & $\checkmark$
& \textbf{24.45} & \textbf{4.35}
& \textbf{2.75} & \textbf{73.4} \\
\hline
\end{tabularx}
\end{table}
\noindent
Table~\ref{tab:ablation_decomp} evaluates the two motion tests. Using $z^A$ alone gives the highest trajectory error, whereas $z^B$ alone tracks more accurately but renders slightly worse. Combining both tests gives the best result. Table~\ref{tab:predseg} replaces the GT labels with predictions 
\begin{table}[!h]
\centering
\caption{\textbf{Motion-decomposition CRCD++ - seq C4.}
$z^A$: instance-level test; $z^B$: boundary-band test.
Removing either leaves the other as the sole criterion.}
\label{tab:ablation_decomp}
\begin{tabular*}{\linewidth}{@{\extracolsep{\fill}}cc|ccc@{}}
\hline
$z^A$ & $z^B$ & PSNR$\uparrow$ & ATE$\downarrow$ & mIoU$\uparrow$\\
\hline
$\checkmark$ & $\times$ & 24.39 & 3.03 & 71.9\\
$\times$ & $\checkmark$ & 24.29 & 2.80 & 72.6\\
\hline
$\checkmark$ & $\checkmark$ & \textbf{24.45} & \textbf{2.75} & \textbf{73.4}\\
\hline
\end{tabular*}
\end{table}
from a pretrained Mask R-CNN~\cite{oh2025expanded} for organs and PFedSIS~\cite{xu2025pfedsis} for tools from previous works, proving that even with estimated segmentations our method outperforms the second-best DDS-SLAM that takes as input ground truth segmentations.

\begin{table}[t]
\centering
\caption{\textbf{Predicted segmentation Avg. performance on CRCD++.}}
\label{tab:predseg}
\setlength{\tabcolsep}{2pt}
\newcolumntype{Y}{>{\centering\arraybackslash}X}
\begin{tabularx}{\textwidth}{l|YYYY}
\hline
Method & PSNR (dB)$\uparrow$ & Depth (mm)$\downarrow$ & ATE (mm)$\downarrow$ & mIoU (\%)$\uparrow$\\
\hline
DDS-SLAM (GT) & 21.55 & 13.45 & 4.34 & --\\
Ours (GT) & \underline{23.55} & \textbf{2.96} & \textbf{3.03} & \textbf{84.2}\\
Ours (Pred.) & \textbf{24.28} & \underline{3.54} & \underline{3.30} & \underline{65.8}\\
\hline
\end{tabularx}
\end{table}

\section{Conclusion}
\label{sec:conclusion}
We introduce \textbf{EndoS$^3$}, a semantic Gaussian splatting SLAM framework that uses instance semantics to guide motion decomposition and deformation modelling, alongside \textbf{CRCD++}, a robotic cholecystectomy benchmark for in-depth semantic SLAM evaluation. \textbf{EndoS$^3$} outperforms SOTA on both CRCD++ and SemSup, demonstrating the value of semantic-awareness. Future work includes addressing surgical image artifacts, primarily low illumination conditions, in semantic SLAM.

\bibliography{sn-bibliography}

\end{document}